\documentclass[runningheads]{llncs}
\usepackage{graphicx}

\usepackage{tikz}
\usepackage{comment}
\usepackage{amsmath,amssymb} 
\usepackage{color}
\usepackage{diagbox}
\usepackage{amsthm}
\usepackage{bm}
\usepackage{algorithmic}
\usepackage{algorithm}
\usepackage{graphicx}
\usepackage{pdfpages}
\usepackage{amssymb}
\usepackage{alltt}
\usepackage{epsfig}
\usepackage{subfigure}
\usepackage{makeidx}
\usepackage{amsmath}
\usepackage{multirow}
\usepackage{mathrsfs}
\usepackage{latexsym}
\usepackage{color}
\usepackage{enumerate}
\usepackage{threeparttable}
\usepackage{wrapfig}
\usepackage{booktabs}
\usepackage{array}
\usepackage{adjustbox}
\usepackage{xspace}
\usepackage{pifont} 
\usepackage[misc]{ifsym} 
\usepackage{hyperref}

\makeatletter
\DeclareRobustCommand\onedot{\futurelet\@let@token\@onedot}
\def\@onedot{\ifx\@let@token.\else.\null\fi\xspace}
\def\eg{\emph{e.g}\onedot} 
\def\ie{\emph{i.e}\onedot}

\def\etal{\emph{et al}\onedot}
\makeatother

\definecolor{darkgreen}{RGB}{0,127,0}
\definecolor{darkred}{RGB}{200,0,0}

\usepackage[accsupp]{axessibility}  

\newcommand{\tf}[1]{{\color{red} [ToFill]}}
\newcommand{\tc}[1]{{\color{red} [ToCite]}}

\makeatletter
\def\blfootnote{\xdef\@thefnmark{}\@footnotetext}
\makeatother

\begin{document}
\begin{sloppypar}
\pagestyle{headings}
\mainmatter

\title{Audio$-$Visual Segmentation} 


\author{$^\star$Jinxing Zhou$^{1,2}$,
        $^\star$Jianyuan Wang$^{3}$,
        Jiayi Zhang$^{2,4}$,
        Weixuan Sun$^{2,3}$,\\
        Jing~Zhang${^{3}}$, 
        Stan Birchfield${^5}$,
        Dan Guo${^1}$,
        Lingpeng Kong$^{6,7}$,\\
        $^\textrm{\Letter} $Meng Wang$^{1}$,
        and $^\textrm{\Letter} $Yiran Zhong$^{2,7}$} 
 \authorrunning{J. Zhou et al.}
 
\institute{
$^{1}$Hefei University of Technology, $^{2}$SenseTime Research,\\ $^{3}$Australian National University, $^{4}$Beihang University, $^{5}$NVIDIA,\\ $^{6}$The University of Hong Kong, $^{7}$Shanghai Artificial Intelligence Laboratory\\
	\email{\{eric.mengwang, zhongyiran\}@gmail.com}}

\maketitle

\begin{abstract}
We propose to explore a new problem called audio-visual segmentation (AVS), in which the goal is to output a pixel-level map of the object(s) that produce sound at the time of the image frame.
To facilitate this research, we construct the first audio-visual segmentation benchmark (AVSBench), providing pixel-wise annotations for the sounding objects in audible videos. 
Two settings are studied with this benchmark: 1) semi-supervised audio-visual segmentation with a single sound source and 2) fully-supervised audio-visual segmentation with multiple sound sources.  
To deal with the AVS problem, we propose a new method that uses a temporal pixel-wise audio-visual interaction module to inject audio semantics as guidance for the visual segmentation process. %
We also design a regularization loss to encourage the audio-visual mapping during training.
Quantitative and qualitative experiments on the AVSBench compare our approach to several existing methods from related tasks, demonstrating that the proposed method is promising for building a bridge between the audio and pixel-wise visual semantics.
Code is available at \color{blue}{\href{https://github.com/OpenNLPLab/AVSBench}{https://github.com/OpenNLPLab/AVSBench}}\blfootnote{\noindent $^{\star}$Equal contribution. $^\textrm{\Letter}$Corresponding author. This work is done when Jinxing Zhou is an intern at SenseTime Research. }.

\end{abstract}


\section{Introduction}
A human can classify an object not only from its visual appearance but also from the sound it makes. For example, when we hear a dog bark or a siren wail, we know the sound is from a dog or ambulance, respectively.  Such observations confirm that the audio and visual information complement each other.

\begin{figure}[t]
\centering
\includegraphics[width=\textwidth]{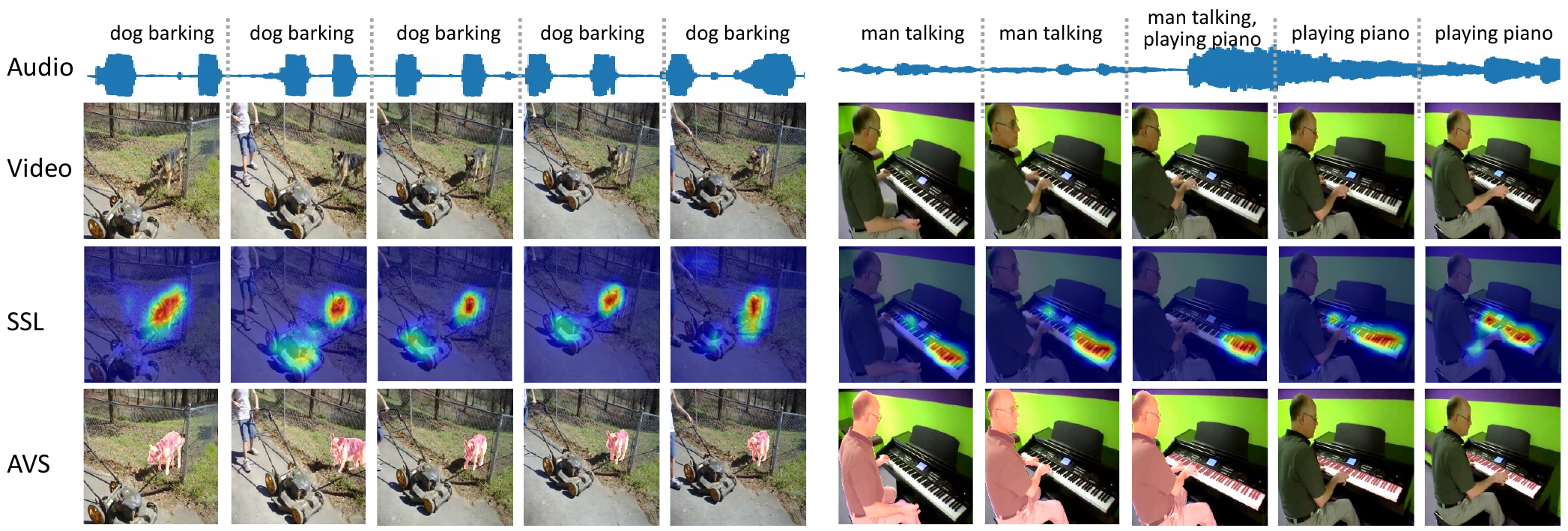}
\vspace{-6mm}
\caption{\textbf{Comparison of the proposed AVS task with the SSL task.} Sound source localization (SSL) estimates a rough location of the sounding objects in the visual frame, at a patch level. We propose AVS to estimate pixel-wise segmentation masks for all the sounding objects, no matter the number of visible sounding objects. {\sc Left:} Video of dog barking. {\sc Right:}  Video with two sound sources (man and piano).}
\label{fig:task_comparison}
\vspace{-5mm}
\end{figure}

To date, researchers have studied this problem from somewhat simplified scenarios.
Some researchers have investigated audio-visual correspondence (AVC)~\cite{arandjelovic2017look,arandjelovic2018objects,aytar2016soundnet} problem, which aims to determine whether an audio signal and a visual image describe the same scene. AVC is based on the phenomenon that these two signals usually occur simultaneously, \eg, a barking dog and a humming car. 
Others studied audio-visual event localization (AVEL)~\cite{lin2019dual,lin2020audiovisual,tian2018audio,wu2019dual,xu2020cross,xuan2020cross,ramaswamy2020makes,ramaswamy2020see,duan2021audio,zhou2021positive}, which classifies the segments of a video into the pre-defined event labels.
Similarly, some people have also explored audio-visual video parsing (AVVP)~\cite{tian2020unified,wu2021exploring,lin2021exploring,yu2021mm}, whose goal is to divide a video into several events and classify them as audible, visible, or both.
Due to a lack of pixel-level annotations, all these scenarios are restricted to the frame/temporal level, thus reducing the problem to that of audible image classification.

A related problem, known as sound source localization (SSL), aims to locate the visual regions within the frames that correspond to the sound~\cite{arandjelovic2017look,arandjelovic2018objects,senocak2018learning,cheng2020look,owens2018audio,chen2021localizing,hu2019deep,qian2020multiple,hu2020discriminative}.
Compared to AVC/AVEL/AVVP, the problem of SSL seeks patch-level scene understanding, \ie, the results are usually presented by a heat map that is obtained either by visualizing the similarity matrix of the audio feature and the visual feature map, or by class activation mapping (CAM)~\cite{zhou2016cam}---without considering the actual shape of the sounding objects.

In this paper, we propose the pixel-level audio-visual segmentation (AVS) problem, which requires the network to densely predict whether each pixel corresponds to the given audio, so that a mask of the sounding object(s) is generated. Fig.~\ref{fig:task_comparison} illustrates the differences between AVS and SSL. The AVS task is more challenging than previous tasks as it requires the network to not only locate the audible frames but also delineate the shape of the sounding objects.

To facilitate this research, we propose AVSBench, the first pixel-level audio-visual segmentation benchmark that provides ground truth labels for sounding objects. 
We divide our AVSBench dataset into two subsets, depending on the number of sounding objects in the video (single- or multi-source).
With AVSBench, we study two settings of audio-visual segmentation:  1) semi-supervised Single Sound Source Segmentation (S4), and 2) fully-supervised Multiple Sound Source Segmentation (MS3). 
For both settings, the goal is to segment the object(s) from the visual frames that are producing sounds.
We test six methods from related tasks on AVSBench and provide a new AVS method as a strong baseline. The latter utilizes a standard encoder-decoder architecture but with a new temporal pixel-wise audio-visual interaction (TPAVI) module to better introduce the audio semantics for guiding visual segmentation. We also propose a loss function to utilize the correlation of audio-visual signals, which further enhances segmentation performance.

Our contributions can be summarized as follows: 1) We propose AVS as a fine-grained audio-visual scene understanding task, and introduce AVSBench, a new dataset providing pixel-level annotations for AVS; 2) We design an end-to-end framework for AVS, which adopts a new TPAVI module to encode temporal pixel-wise audio-visual interaction, and a regularization loss to further utilize the audio-visual correlation; and 3) We conduct extensive experiments to verify the benefits of considering audio signals for visual segmentation.  We also compare with several relevant methods to show the superiority of our proposed approach in both settings.

\vspace{-3mm}
\section{Related Work}\label{sec:related_work}
\textbf{Sound Source Localization (SSL).} The most closely related problem to ours is SSL, which aims to locate the regions in the visual frames responsible for the sounds. The prediction of SSL is usually computed from the similarity matrix of the learned audio feature and the visual feature map~\cite{arandjelovic2017look,arandjelovic2018objects,senocak2018learning,cheng2020look,owens2018audio,chen2021localizing}, displayed as a heat map. 
SSL can also be divided into two settings according to the complexity of sound sources, \emph{viz.}, single and multiple sound source(s) localization. Here we focus on the challenging setting of multiple sources, which requires accurately localizing the true sound source among multiple potential candidates~\cite{hu2019deep,afouras2020self,qian2020multiple,hu2020discriminative}. 
In pioneering work, Hu \etal~\cite{hu2019deep} divide the audio and visual features into multiple cluster centers and take the center distance as a supervision signal to rank the paired audio-visual information.
Qian \etal ~\cite{qian2020multiple} first train an audio-visual correspondence model to extract coarse feature representations of audio and visual signals, and then use Grad-CAM~\cite{selvaraju2017grad,sun2021getam} to visualize the class-specific features for localization. 
Furthermore, Hu \etal~\cite{hu2020discriminative} adopt a two-stage method, which first learns audio-visual semantics in the single sound source condition, using such learned knowledge to help with multiple sound sources localization.
Rouditchenko \etal ~\cite{rouditchenko2019self} tackle this problem by disentangling category concepts in the neural networks.
This method is actually more related to the task of \emph{sound source separation}~\cite{zhao2018sound,gao2018learning,zhao2019sound,gao2019co} and shows sub-optimal performance regarding visual localization.
Although these existing SSL methods indicate which regions in the image are making sound, the results do not clearly delineate the shape of the objects. 
Rather, the location map is computed by up-sampling the audio-visual similarity matrix from a low resolution.
Moreover, the methods above all rely on unsupervised learning when capturing the shape of sounding objects, which partly suffers from the lack of an annotated dataset.
To overcome these limitations, this paper provides an audio-visual segmentation dataset with pixel-level ground truth labels, which enables to achieve more accurate segmentation predictions.

\noindent\textbf{Audio-Visual Dataset.}
To the best of our knowledge, there are no publicly available datasets that provide segmentation masks for the sounding visual objects with audio signals.
Here we briefly introduce the popular datasets in the audio-visual community.
For example, the AVE~\cite{tian2018audio} and LLP~\cite{tian2020unified} datasets are respectively collected for audio-visual event localization and video parsing tasks. 
They only have category annotations for video frames, and hence cannot be used for pixel-level segmentation. 
For the sound source localization problem, researchers usually use the Flickr-SoundNet~\cite{senocak2018learning} and VGG-SS~\cite{chen2021localizing} datasets, where the videos are sampled from the large-scale Flickr~\cite{aytar2016soundnet} and VGGSound~\cite{chen2020vggsound} datasets, respectively.
The authors provide bounding boxes to outline the location of the target sound source, which could serve as patch-level supervision. 
%
%
However, this still inevitably suffers from incorrect evaluation results since the sounding objects are usually irregular in shape and some regions within the bounding box actually do not belong to the real sound source. 
%
%
This is a reason why current sound source localization methods can only roughly locate sounding objects but cannot learn their accurate shapes, which inhibits the mapping from audio signals to fine-grained visual cues.

\section{The AVSBench}\label{sec:dataset}

\begin{figure}[t]
\centering
\includegraphics[width=0.95\textwidth]{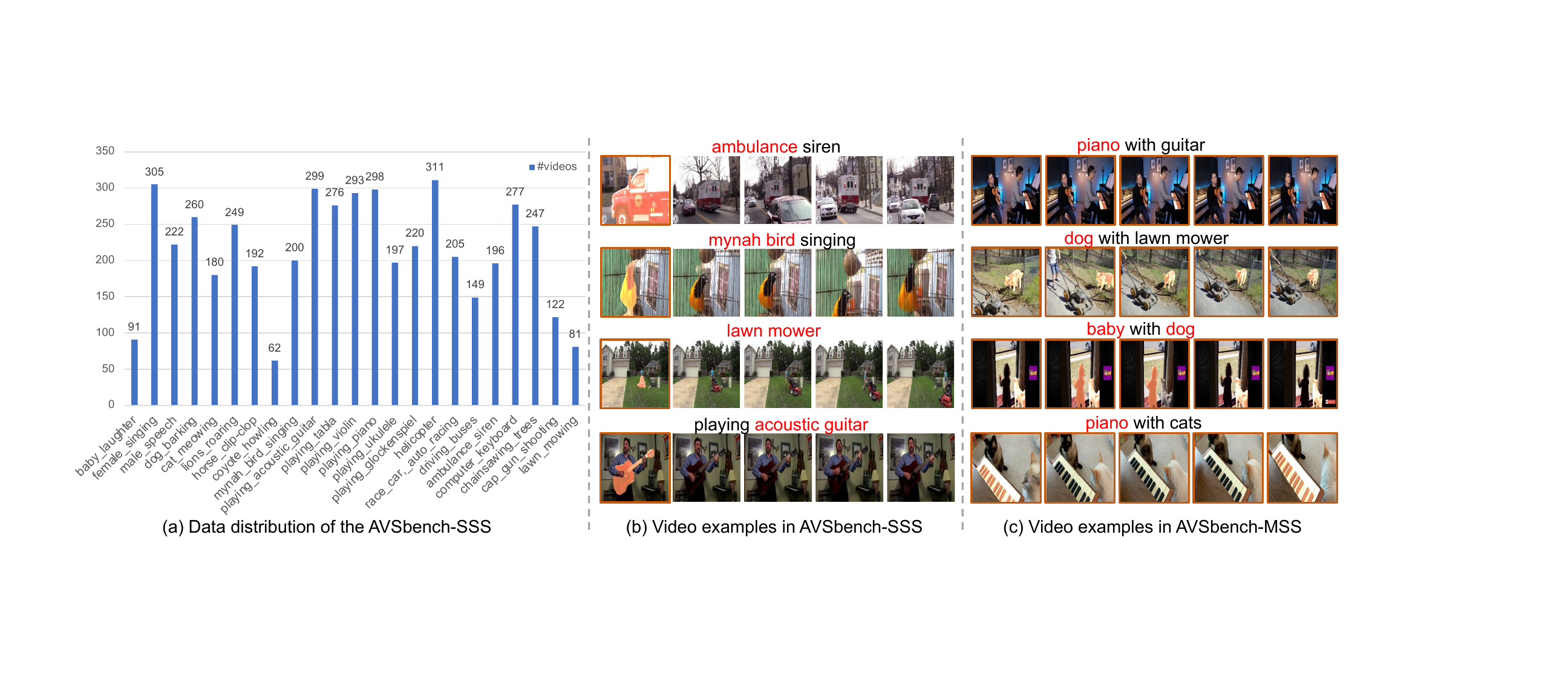}
\caption{\textbf{Statistics of the whole Single-source subset of AVSBench.} The texts represent the category names. For example, the `helicopter' category contains $311$ video samples. }
\label{fig:single_source_statistics}
\end{figure}

\subsection{Dataset Statistics} 
\begin{figure}[t]
\centering
\includegraphics[width=\textwidth]{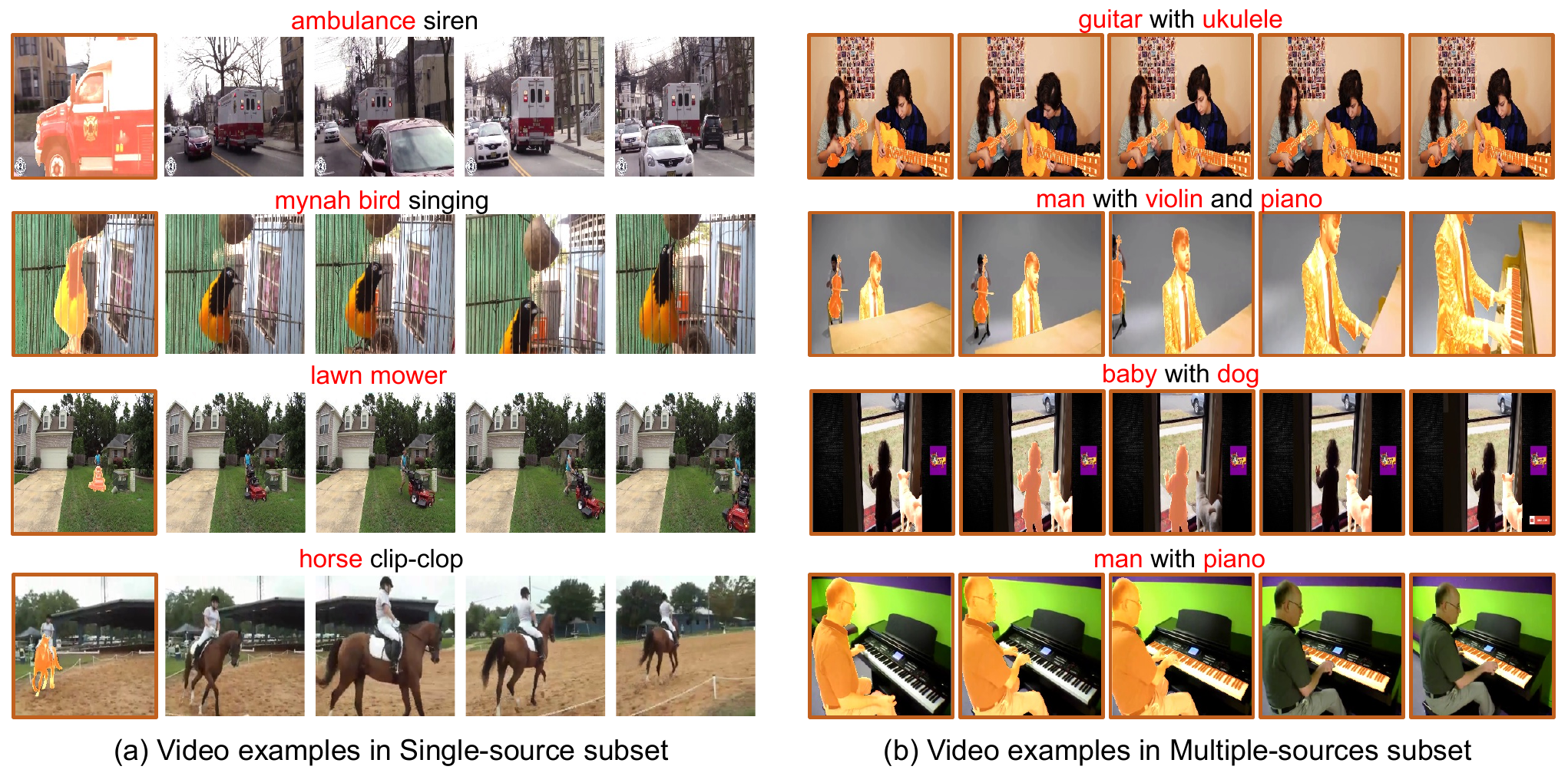}
\vspace{-4mm}
\caption{\textbf{AVSBench samples}. The AVSBench dataset contains the Single-source subset ({\sc Left}) and Multi-sources subset ({\sc Right}).
Each video is divided into 5 clips, as shown.
Annotated clips are indicated by brown framing rectangles; the name of sounding objects is indicated by red text.
Note that for Single-source training set, only the first frame of each video is annotated, whereas 5 frames are annotated for all other sets.
}
\label{fig:AVSbench_examples}
\vspace{-6mm}
\end{figure}

AVSBench is designed for pixel-level audio-visual segmentation. We collected the videos using the techniques introduced in VGGSound~\cite{chen2020vggsound} to ensure the audio and visual clips correspond to the intended semantics. AVSBench contains two subsets---Single-source and Multi-sources---depending on the number of sounding objects. All videos were downloaded from YouTube with the \textit{Creative Commons} license, and each video was trimmed to $5$ seconds. The Single-source subset contains $4,932$ videos over $23$ categories, covering sounds from humans, animals, vehicles, and musical instruments. 
In Fig.~\ref{fig:single_source_statistics}, we show the category names and the video number for each category.
To collect the Multi-sources subset, we selected the videos that contain multiple sounding objects, \eg, a video of baby laughing, man speaking, and then woman singing. To be specific, we randomly chose two or three category names from the Single-source subset as keywords to search for online videos, then manually filtered out videos to ensure 1) each video has multiple sound sources, 2) the sounding objects are visible in the frames, and 3) there is no deceptive sound, \eg, canned laughter. In total, this process yielded $424$ videos for the Multi-sources subset out of more than six thousand candidates.
The ratio of train/validation/test split percentages are set as 70/15/15 for both subsets, as shown in Table~\ref{table:dataset_split}. 
Several video examples are visualized in Fig.~\ref{fig:AVSbench_examples}, where the red text indicates the name of sounding objects.

\begin{table}[t]
\begin{center}
\caption{\textbf{AVSBench statistics}. The videos are split into train/valid/test.  The asterisk ($^*$) indicates that, for Single-source training, one annotation per video is provided; all others contain 5 annotations per video.  (Since there are 5 clips per video, this is 1 annotation per clip.)  Together, these yield the total annotated frames.}
\label{table:dataset_split}
\setlength{\tabcolsep}{6pt}
\begin{tabular}{ccccc}
\toprule[0.8pt]\noalign{\smallskip}
subset & classes & videos & train/valid/test  & annotated frames \\
\noalign{\smallskip}
\midrule
\noalign{\smallskip}
Single-source & 23 & 4,932 & 3,452$^*$/740/740 & 10,852\\
Multi-sources & 23  & \,\,\,424   & 296/64/64 & \,\,\,2,120 \\
\bottomrule[0.8pt]
\end{tabular}
\end{center}
\vspace{-5mm}
\end{table}

\begin{table}[t]
\begin{center}
\caption{\textbf{Existing audio-visual dataset statistics.}
Each benchmark is shown with the number of videos and the \emph{annotated} frames. 
The final column indicates whether the frames are labeled by category, bounding boxes, or pixel-level masks.}
\label{table:comparison_with_datasets}
\setlength{\tabcolsep}{6pt}
\begin{tabular}{cccccc}
\toprule[0.8pt]\noalign{\smallskip}
benchmark & videos & frames & classes & types  &  annotations \\
\noalign{\smallskip}
\hline
AVE  \cite {tian2018audio}           & \,\,\,4,143 & 41,430  & \,\,\,28  & video  & category   \\ 
LLP \cite{tian2020unified}             & 11,849 & 11,849 & \,\,\,25  & video & category  \\
Flickr-SoundNet \cite{senocak2018learning} & \,\,\,5,000 & \,\,\,5,000  & \,\,\,50  & image & bbox   \\
VGG-SS \cite{chen2021localizing}         & \,\,\,5,158 & \,\,\,5,158  & 220  & image & bbox    \\  
AVSBench (ours)        & \,\,\,5,356 & 12,972 & \,\,\,23   & video & pixel    \\
\bottomrule[0.8pt]
\end{tabular}
\end{center}
\vspace{-8mm}
\end{table}

In addition, we make a comparison between AVSBench with other popular audio-visual benchmarks in Table~\ref{table:comparison_with_datasets}.
The AVE~\cite{tian2018audio} dataset contains 4,143 videos covering 28 event categories. 
The LLP~\cite{tian2020unified} dataset consists of 11,849 YouTube video clips spanning with $25$ categories, collected from AudioSet~\cite{gemmeke2017audio}.
Both the AVE and LLP datasets are labelled at a frame level, through audio-visual event boundaries.
Meanwhile, the Flickr-SoundNet~\cite{senocak2018learning} dataset and VGG-SS~\cite{chen2021localizing} dataset are proposed for sound source localization (SSL), labelled at a patch level through bounding boxes.
In contrast, our AVSBench contains 5,356 videos with 12,972 pixel-wise annotated frames. The benchmark is designed to facilitate research on fine-grained audio-visual segmentation. Additionally, it provides accurate ground truth for sound source localization, which could help the training of SSL methods and serve as an evaluation benchmark for that problem as well.

\subsection{Annotation} We divide each 5-second video into five equal 1-second clips, and we provide manual pixel-level annotations for the last frame of each clip. For this sampled frame, the ground truth label is a binary mask indicating the pixels of sounding objects, according to the audio at the corresponding time. For example, in the Multi-sources subset, even though a dancing person shows drastic movement spatially, it would not be labelled as long as no sound was made. 
In clips where objects do not make sound, the object should not be masked, \eg, the \emph{piano} in the first two clips of the last row of Fig.~\ref{fig:AVSbench_examples}b. Similarly, when more than one object emits sound, all the emitting objects are annotated, \eg, the guitar and ukulele in the
first row in Fig.~\ref{fig:AVSbench_examples}b.
Also, when the sounding objects in the video are dynamically changing,
the difficulty is further increased, \eg, the second, third, and fourth rows in Fig.~\ref{fig:AVSbench_examples}b.
Currently, for large-scale objects, we only annotate their most representative parts.
For example, we label the keyboard of pianos because it is sufficiently recognizable, while, the cabinet part of pianos is often too varied.

We use two types of labeling strategies, based on the different difficulties between the Single-source and the Multi-sources subsets. For the videos in the training split of Single-source, we only annotate the first sampled frame (with the assumption that the information from one-shot annotation is sufficient, as the Single-source subset has a single and consistent sounding object over time). This assumption is verified by the quantitative experimental results shown in Table~\ref{table:comparison_with_baselines}.
For the more challenging Multi-sources subset, all clips are annotated for training, since the sounding objects may change over time. 
Note that for validation and test splits, all clips are annotated, as shown in Table~\ref{table:dataset_split}.

\begin{figure}[t]
\centering
\includegraphics[width=\textwidth]{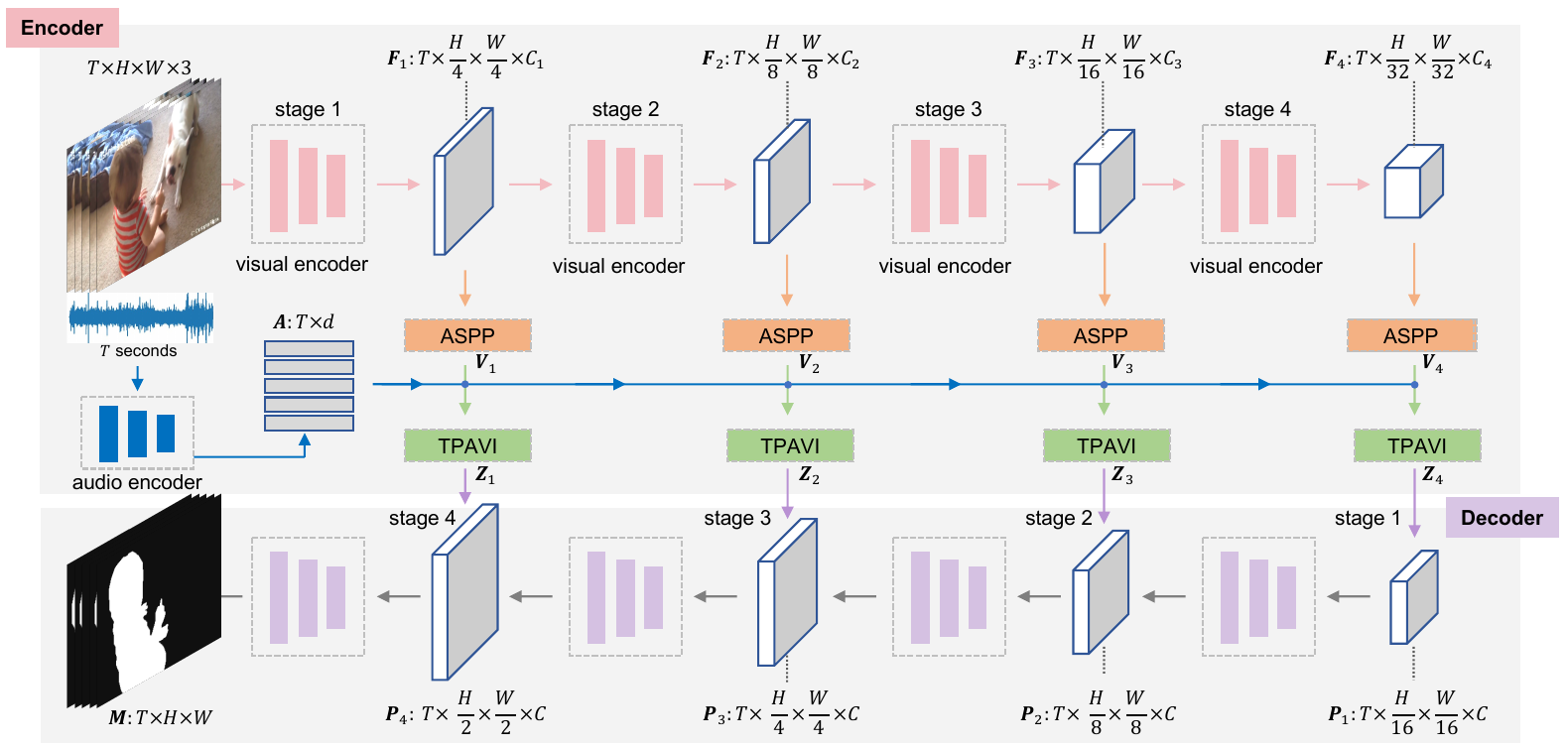}
\caption{\textbf{Overview of the Baseline}, which follows a hierarchical {Encoder-Decoder} pipeline. The \emph{encoder} takes the video frames and the entire audio clip as inputs, and outputs visual and audio features, respectively denoted as $\bm{F}_i$ and $\bm{A}$.  The visual feature map $\bm{F}_i$ at each stage is further sent to the ASPP~\cite{chen2017deeplab} module and then our TPAVI module (introduced in Sec.~\ref{sec:approach}). ASPP provides different receptive fields for recognizing visual objects, while TPAVI focuses on the temporal pixel-wise audio-visual interaction. The \emph{decoder} progressively enlarges the fused feature maps by four stages and finally generates the output mask $\bm{M}$ for sounding objects.}
\vspace{-3mm}
\label{fig:framework}
\end{figure}

\subsection{Two Benchmark Settings}\label{sec:problem_statement}
We provide two benchmark settings for our AVSBench dataset: the semi-supervised Single Sound Source Segmentation (S4) and the fully supervised Multiple Sound Source Segmentation (MS3). For ease of expression, we denote the video sequence as $S$, which consists of $T$ non-overlapping yet continuous clips $\{S_t^v, S_t^a\}_{t=1}^{T}$, where $S^v$ and $S^a$ are the visual and audio components, and $T=5$. In practice, we extract the video frame at the end of each second.

\noindent\textbf{Semi-supervised S4} corresponds to the Single-source subset. It is termed as semi-supervised because only part of the ground truth is given during training (\ie, the first sampled frame of the videos) but all the video frames require a prediction during evaluation. We denote the pixel-wise label as $\bm{Y}_{t=1}^s \in \mathbb{R}^{H \times W}$, where $H$ and $W$ are the frame height and width, respectively. $\bm{Y}_{t=1}^s$ is a binary matrix where $1$ indicates sounding objects while $0$ corresponds to background or silent objects. 

\noindent\textbf{Fully-supervised MS3} deals with the Multi-sources subset, where the labels of all five sampled frames of each video are available for training. The ground truth is denoted as $\{\bm{Y}_t^m\}_{t=1}^{T}$, where $\bm{Y}_t^{m} \in \mathbb{R}^{H \times W}$ is the binary label for the $t$-th video clip.

The goal for both settings is to correctly segment the sounding object(s) for each video clip by utilizing the audio and visual cues, \ie, $S^a$ and $S^v$. Generally, it is expected $S^a$ to indicate the target object, while $S^v$ provides information for fine-grained segmentation. 
The predictions are denoted as $\{\bm{M}_t\}_{t=1}^{T}$, $\bm{M}_t \in \mathbb{R}^{H \times W}$. Both the semi-supervised and fully-supervised settings are conducted in a category-agnostic way such that the models work for general videos.

\section{A Baseline}\label{sec:approach}

We propose a new baseline method for the pixel-level audio-visual segmentation (AVS) task as shown in Fig.~\ref{fig:framework}. We use the same framework in both semi- and fully-supervised settings. Following the convention of semantic segmentation methods \cite{jon2014fcn,ron2015unet,Wang2021PVTv2IB,xie2021segformer}, our method adopts an encoder--decoder architecture.

\noindent\textbf{The Encoder:} We extract audio and visual features independently. Given an audio clip $S^a$, we first process it to a spectrogram via the short-time Fourier transform, and then send it to a convolutional neural network, VGGish~\cite{hershey2017cnn}. We use the weights that are pretrained on AudioSet~\cite{gemmeke2017audio} to extract audio features $\bm{A} \in \mathbb{R}^{T \times d}$, where $d=128$ is the feature dimension.
For a video frame $S^v$, we extract visual features with popular convolution-based or vision transformer-based backbones. We try both two options in the experiments and they show similar performance trends. These backbones produce hierarchical visual feature maps during the encoding process, as shown in Fig.~\ref{fig:framework}. 
We denote the features as $\bm{F}_i \in \mathbb{R}^{T \times h_i \times w_i \times C_i}$, where $(h_i,w_i)=(H,W)/2^{i+1}$, $i=1,\ldots,n$. 
The number of levels is set to $n=4$ in all experiments.

\begin{wrapfigure}[19]{r}{0.45\textwidth}
\vspace{-13mm}
\begin{center}
\includegraphics[width=0.45\textwidth]{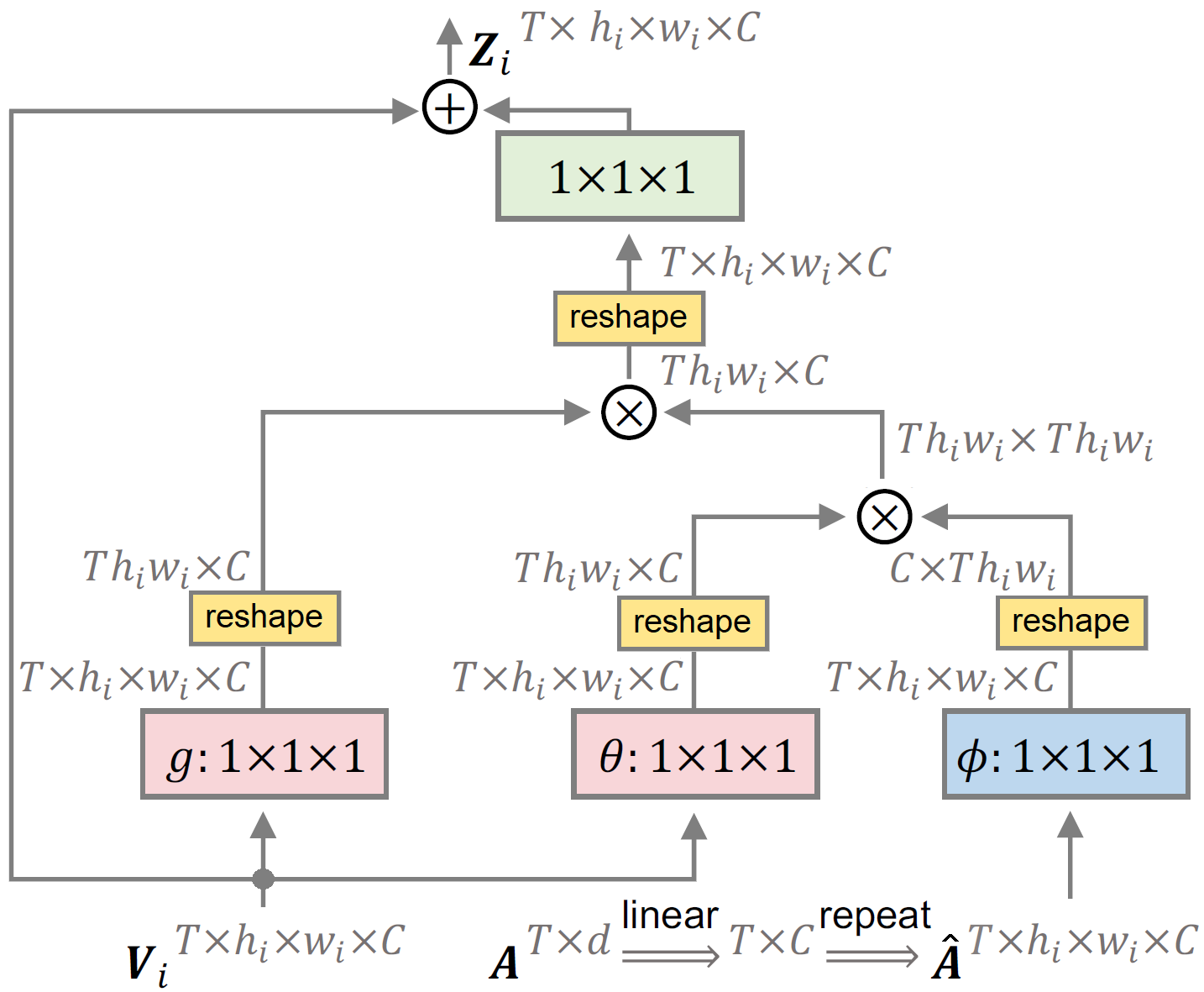}
\vspace{-8mm}
  \caption{\textbf{The TPAVI module} takes the $i$-th stage visual feature $\bm{V}_i$ and the audio feature $\bm{A}$ as inputs. The colored boxes represent $1 \times 1 \times 1$ convolutions, while the yellow boxes indicate reshaping operations. The symbols ``$\otimes$'' and ``$\oplus$'' denote matrix multiplication and element-wise addition, respectively.}
  \label{fig:TPAVI_module}
  \end{center}
\end{wrapfigure}

\noindent\textbf{Cross-Modal Fusion:} We use Atrous Spatial Pyramid Pooling (ASPP) modules~\cite{chen2017deeplab} to further post-process the visual features $\bm{F}_i$ to $\bm{V}_i \in \mathbb{R}^{T \times h_i \times w_i \times C}$, where $C=256$. These modules employ multiple parallel filters with different rates and hence help to recognize visual objects with different receptive fields, \eg, different sized moving objects. Then, we consider introducing the audio information to build the audio-visual mapping to assist with identifying the sounding object. This is particularly essential for the MS3 setting where there are multiple dynamic sound sources.
Our intuition is that, although the auditory and visual signals of the sound sources may not appear simultaneously, they usually exist in more than one video frame.
Therefore, integrating the audio and visual signals of the whole video should be beneficial. 
Motivated by~\cite{wang2018non} that uses the non-local block to encode space-time relation, we adopt a similar module to encode the temporal pixel-wise audio-visual interaction (TPAVI).
As illustrated in Fig.~\ref{fig:TPAVI_module}, the current visual feature map $\bm{V}_i$ and the audio feature $\bm{A}$ of the entire video are sent into the TPAVI module.
Specifically, the audio feature $\bm{A}$ is first transformed to a feature space with the same dimension as the visual feature $\bm{V}_i$, by a linear layer.
Then it is spatially duplicated $h_i w_i$ times and reshaped to the same size as $\bm{V}_i$. We denote such processed audio feature as $\hat{\bm{A}}$. Next, it is expected to find those pixels of visual feature map $\bm{V}_i$ that have a high response to the audio counterpart $\hat{\bm{A}}$ through the entire video.

Such an audio-visual interaction can be measured by dot-product, then the updated feature maps $\bm{Z}_i$ at the $i$-th stage can be computed as,
\begin{equation}\label{eq:tpavi}
\footnotesize
\textstyle
\begin{aligned}
    & \bm{Z}_i = \bm{V}_i + \mu(\alpha_i\ {g(\bm{V}_i})), \ \text{where} \ \alpha_i = \frac{\theta(\bm{V}_i) \ {\phi(\hat{\bm{A}})}^{\top}}{N} \\
\end{aligned}
\end{equation}

\noindent where $\theta$, $\phi$, $g$ and $\mu$ are  $1 \times 1 \times 1$ convolutions, $N=T \times h_i \times w_i$ is a normalization factor, $\alpha_i$ denotes the audio-visual similarity, and $\bm{Z}_i \in \mathbb{R}^{T \times h_i \times w_i \times C}$. Each visual pixel interacts with all the audios through the TPAVI module. 
 We provide a visualization of the audio-visual attention in TPAVI later in Fig.~\ref{fig:audio_visual_attention}, which shows a similar ``appearance'' to the prediction of SSL methods because it constructs a pixel to audio mapping.

\noindent\textbf{The Decoder:} We adopt the decoder of Panoptic-FPN~\cite{kirillov2019panoptic} in this work for its flexibility and effectiveness, though any valid decoder architecture could be used. In short, at the $j$-th stage, where $j=2,3,4$, both the outputs from stage $\bm{Z}_{5-j}$ and the last stage $\bm{Z}_{6-j}$ of the encoder are utilized for the decoding process. The decoded features are then upsampled to the next stage. The final output of the decoder is $\bm{M} \in \mathbb{R}^{T \times H \times W}$, activated by \emph{sigmoid}.


\noindent\textbf{Objective function:}\label{sec:loss_function}
Given the prediction $\bm{M}$ and the pixel-wise label $\bm{Y}$, we adapt the binary cross entropy (BCE) loss as the main supervision function. Besides, we use an additional regularization term $\mathcal{L}_\text{AVM}$ to force the audio-visual mapping. Specifically, we use the Kullback–Leibler (KL) divergence to ensure the masked visual features have similar distributions with the corresponding audio features. In other words, if the audio features of some frames are close in feature space, the corresponding sounding objects are expected to be close in feature space. The total objective function $\mathcal{L}$ can be computed as follows:
\begin{align}\label{eq:loss}
  &\mathcal{L} = \text{BCE}(\bm{M}, \bm{Y}) + \lambda \mathcal{L}_\text{AVM}(\bm{M},\bm{Z},\bm{A}), \\
  &\mathcal{L}_\text{AVM} = \sum_{i=1}^{n}(\text{KL}( avg \ (\bm{M}_i \odot \bm{Z}_i), \bm{A}_i)), \label{eq:avmloss}
\end{align}
where $\lambda$ is a balance weight, $\odot$ denotes element-wise multiplication, and $\textit{avg}$ denotes the average pooling operation. 
At each stage, we down-sample the prediction $\bm{M}$ to $\bm{M}_i$ via average pooling to have the same shape as $\bm{Z}_i$.
The vector $\bm{A}_i$ is a linear transformation of $\bm{A}$ that has the same feature dimension with $\bm{Z}_i$.
For the semi-supervised S4 setting, we found that the audio-visual regularization loss does not help, so we set $\lambda=0$ in this setting.

\section{Experimental Results}
\subsection{Implementation details}\label{sec:experimental_setup}

We conduct training and evaluation on the proposed AVSBench dataset, with both convolution-based and transformer-based backbones, ResNet-50~\cite{he2016deep} and Pyramid Vision Transformer~(PVT-v2)~\cite{Wang2021PVTv2IB}.
The backbones have been pretrained on the ImageNet~\cite{russakovsky2015imagenet} dataset.
All the video frames are resized to the shape of $224 \times 224$. 
The channel sizes of the four stages are $C_{1:4} = [256, 512, 1024, 2048]$ and $C_{1:4} = [64, 128, 320, 512]$ for ResNet-50 and PVT-v2, respectively. 
The channel size of the ASPP module is set to $C=256$.
We use the VGGish model to extract audio features, a VGG-like network~\cite{hershey2017cnn} pretrained on the AudioSet~\cite{gemmeke2017audio} dataset.
The audio signals are converted to one-second splits as the network inputs.
We use the Adam optimizer with a learning rate of 1e-4 for training.
The batch size is set to $4$ and the number of training epochs are $15$ and $30$ respectively for the semi-supervised S4 and the fully-supervised MS3 settings. 
The $\lambda$ in Eq.~\eqref{eq:loss} is empirically set to 0.5. 

\subsection{Comparison with methods from related tasks}\label{sec:exp_compare_to_ssl}
We compare our baseline framework with the methods from three related tasks, including sound source localization (SSL), video object segmentation (VOS), and salient object detection (SOD). 
For each task, we report the results of two SOTA methods on our AVSBench dataset, \ie, LVS~\cite{chen2021localizing} and MSSL~\cite{qian2020multiple} for SSL, 3DC~\cite{mahadevan2020making} and SST~\cite{duke2021sstvos} for VOS, iGAN~\cite{mao2021transformer} and LGVT~\cite{zhang2021learning} for SOD. We select these methods as they are the state-of-the-art in their fields: 1) \textit{LVS} uses the background and the most confident regions of sounding objects to design a contrastive loss for audio-visual representation learning and the localization map is obtained by computing the audio-visual similarity.
2) \textit{MSSL} is a two-stage method for multiple sound source localization and the localization map is obtained by Grad-CAM~\cite{selvaraju2017grad}.
3) \textit{3DC} adopts an architecture that is fully constructed by powerful 3D convolutions to encode video frames and predict segmentation masks.
4) \textit{SST} introduces a transformer architecture to achieve sparse attention of the features in the spatiotemporal domain.
5) \textit{iGAN} is a ResNet-based generative model for saliency detection, considering about the inherent uncertainty of saliency detection.
6) \textit{LGVT} is a saliency detection method based on Swin transformer \cite{liu2021swin}, whose long-range dependency modeling ability  leads to a better global context modeling.
We adopt the architecture of these methods and fit them to our semi-supervised S4 and fully-supervised MS3 settings. 
For a fair comparison, the backbones of these methods are all pretrained on the ImageNet~\cite{russakovsky2015imagenet}. 

\begin{table}[t]
\scriptsize
\caption{\textbf{Comparison with methods from related tasks.} Results of the evaluation metrics $\mathcal{M}_{\mathcal{J}}$ and $\mathcal{M}_{\mathcal{F}}$ under both S4 and MS3 settings are reported.}
\begin{center}
 \begin{threeparttable}
  \begin{tabular}{cp{1.1cm}<{\centering}p{1.0cm}<{\centering}p{1.1cm}<{\centering}p{1.1cm}<{\centering}p{1.1cm}<{\centering}p{1.1cm}<{\centering}p{1.1cm}<{\centering}p{1.2cm}<{\centering}p{1.2cm}<{\centering}p{1.1cm}<{\centering}}
  \toprule[0.8pt]
      \multirow{2}{*}{Metric} & \multirow{2}{*}{Setting}   & \multicolumn{2}{c}{SSL}  & \multicolumn{2}{c}{VOS} & \multicolumn{2}{c}{SOD} & \multicolumn{2}{c}{AVS~(ours)}\\ 
    \cmidrule(r){3-4}\cmidrule(r){5-6}\cmidrule(r){7-8}\cmidrule(r){9-10}
                   & & LVS\cite{chen2021localizing} & MSSL\cite{qian2020multiple} & 3DC\cite{mahadevan2020making} & SST\cite{duke2021sstvos} & iGAN\cite{mao2021transformer} & LGVT\cite{zhang2021learning} & ResNet50 & PVT-v2 \\ \midrule
     \multirow{2}{*}{$\mathcal{M}_{\mathcal{J}}$} & S4 & .379 & .449 & .571 & .663 & .616 & .749 & .728 & \textbf{.787}  \\
    & MS3  & .295 & .261 & .369 & .426  & .429 & .407 & .479 & \textbf{.540}  \\ \midrule
     \multirow{2}{*}{$\mathcal{M}_{\mathcal{F}}$} & S4 & .510 & .663 & .759 & .801 & .778 & .873 & .848 & \textbf{.879}  \\
    & MS3  & .330 & .363 & .503  &   .572& .544 & .593 &  .578 & \textbf{.645}  \\
    \bottomrule[0.8pt]
  \end{tabular}
\end{threeparttable}
\end{center}\label{table:comparison_with_baselines}
\vspace{-8mm}
\end{table}

\begin{figure}[t]
\centering
\includegraphics[width=\textwidth]{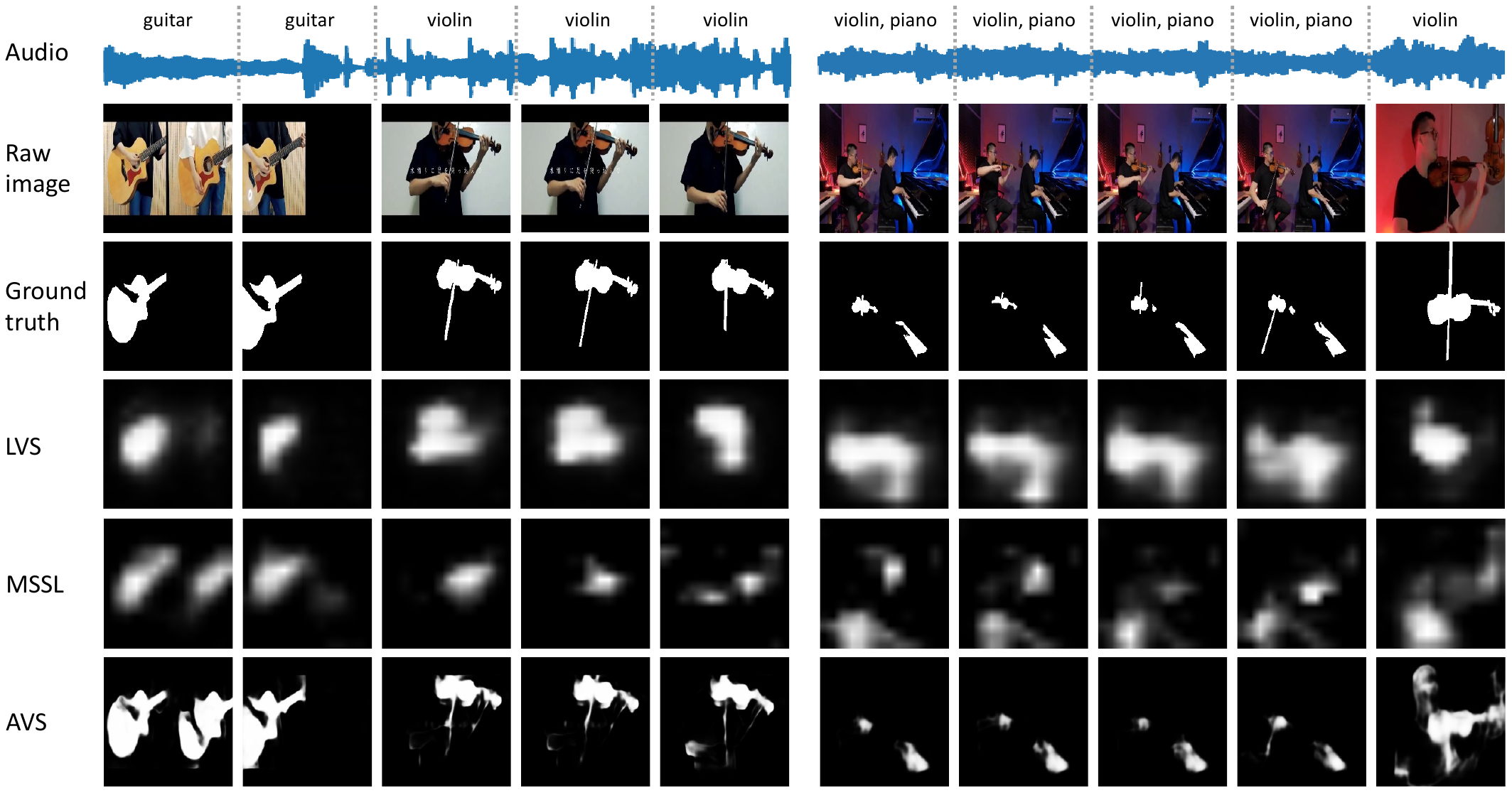}
\caption{\textbf{Qualitative examples of the SSL methods and our AVS framework,} under the fully-supervised MS3 setting. The SSL methods (LVS \cite{chen2021localizing} and MSSL \cite{qian2020multiple}) can only generate rough location maps, while the AVS framework can accurately segment the pixels of sounding objects and nicely outline their shapes.}
\label{fig:example_avs_ssl}
\vspace{-3mm}
\end{figure}

\begin{figure}[t]
\centering
\includegraphics[width=\textwidth]{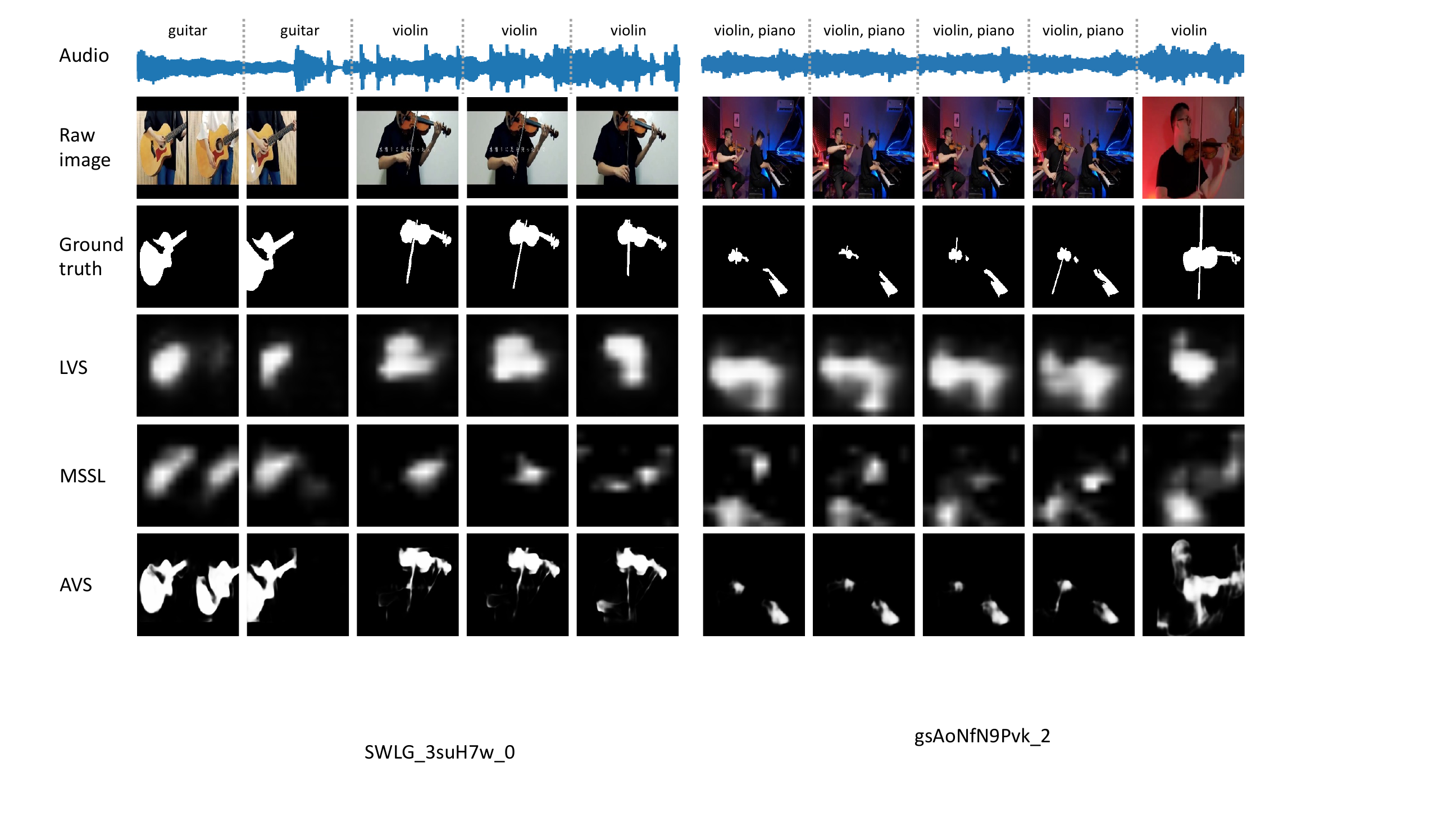}
\vspace{-3mm}
\caption{\textbf{Qualitative examples of the VOS, SOD, and our AVS methods,} under the fully-supervised MS3 setting. We pick the state-of-the-art VOS method SST~\cite{duke2021sstvos} and SOD method LGVT~\cite{zhang2021learning}. As can be verified in the left sample, SST or LGVT cannot capture the change of sounding objects (from `baby' to `baby and dog'), while the AVS accurately conducts prediction under the guidance of the audio signal.
}
\label{fig:example_avs_sod}
\vspace{-3mm}
\end{figure}

\noindent \textbf{Quantitative comparison between AVS and SSL/VOS/SOD.}
We use the Jaccard index $\mathcal{J}$~\cite{everingham2010pascal} and F-score $\mathcal{F}$ as the evaluation metrics\footnote{
$\mathcal{J}$ computes the intersection-over-union of the predicted segmentation and the ground truth mask.
$\mathcal{F}$ considers both the \textsf{precision} and \textsf{recall}: $
    \mathcal{F}_\beta = \frac{(1+\beta^2) \times \mathsf{precision} \times \mathsf{recall}}{\beta^2 \times \mathsf{precision} + \mathsf{recall}}
$, where $\beta^2$ is set to 0.3 in our experiments.
}, where $\mathcal{J}$ and $\mathcal{F}$ measure the region similarity and contour accuracy, respectively.
The quantitative results are shown in Table~\ref{table:comparison_with_baselines}, where $\mathcal{M}_{\mathcal{J}}$ and $\mathcal{M}_{\mathcal{F}}$ denotes the \textit{mean} metric values over the whole dataset.
There is a substantial gap between the results of SSL methods and those of our baseline, mainly because the SSL methods cannot provide pixel-level prediction.
Also, our baseline framework shows a consistent superiority to the VOS and SOD methods in both semi-supervised S4 and fully-supervised MS3 settings.
%
It is worth noting that the state-of-the-art SOD method LGVT~\cite{zhang2021learning} slightly outperforms our ResNet50-based baseline under the Single-source set ($\mathcal{M}_{\mathcal{J}}$: 0.749 \textit{vs.} 0.728), mainly because LGVT uses the strong Swin Transformer backbone~\cite{liu2021swin}.
However, when it comes to the Multi-sources setting, the performance of LGVT is obviously worse than that of our ResNet50-based baseline ($\mathcal{M}_{\mathcal{J}}$: 0.407 \textit{vs.} 0.479).
This is because the SOD method relies on the dataset prior, and cannot handle the situations where sounding objects change but visual contents remain the same (as shown in the left example of Fig.~\ref{fig:example_avs_sod}).
Instead, the audio signals guide our AVS method to identify which object to segment, leading to better performance.
Moreover, if also using a transformer-based backbone, our method is stronger than LGVT in both settings.
Besides, we notice that although SSL methods utilize both the audio and visual signals, they cannot match the performance of VOS or SOD methods that only use visual frames.
It indicates the significance of pixel-wise scene understanding. 
%
The proposed AVS baselines achieve satisfactory performance under the semi-supervised S4 setting ($\mathcal{M}_{\mathcal{J}}$ is around 0.7), which verifies that one-shot annotation is sufficient for single-source case.

\noindent\textbf{Qualitative comparison between AVS and SSL/VOS/SOD.}
We provide some qualitative examples to compare our AVS framework with the SSL methods, LVS~\cite{chen2021localizing} and MSSL~\cite{qian2020multiple}. As shown in the left sample of Fig.~\ref{fig:example_avs_ssl},
LVS over-locates the sounding object \emph{violin}.
At the same time, MSSL fails to locate the \emph{piano} of the right sample.
Both the results of these two methods are blurry and they cannot accurately locate the sounding objects.
Instead, the proposed AVS framework can not only accurately segment all the sounding objects, but also nicely outline the object shapes.

Besides, we also compare the proposed AVS framework with the state-of-the-art methods from VOS and SOD, \ie, SST~\cite{duke2021sstvos} and LGVT~\cite{zhang2021learning}, respectively.
As shown in Fig.~\ref{fig:example_avs_sod},  SST and LGVT can predict their objects of interest in a pixel-wise manner.
However, their predictions rely on the visual saliency and the dataset prior, which cannot satisfy our problem setting. 
For example, in the left sample of Fig.~\ref{fig:example_avs_sod}, the \emph{dog} keeps quiet in the first two frames and should not be viewed as an object of interest in our problem setting.
Our AVS method correctly follows the guidance of the audio signal, \ie, accurately segmenting the \emph{baby} at the first two frames and both the sounding objects at the last three frames, with their shapes complete.
Instead, the VOS method SST misses the barking dog at the last three frames.
The SOD method LGVT masks out both the \emph{baby} and \emph{dog} over all the frames mainly because these two objects usually tend to be `salient', which is not desired in this sample.
When it comes to the right sample of Fig.~\ref{fig:example_avs_sod}, we can observe that LGVT almost fails to capture the \emph{marimba}, since the marimba is in the bottom left corner.
The VOS method SST can find the rough location of the piano but incorrectly segments the man that does not make sounds.
In contrast, our AVS framework can accurately locate all the sounding objects in the first four frames and successfully generate black segmentation map for the last frame as there are no sounding objects.

\noindent\textbf{Comparison with a two-stage baseline.} The AVS task can be tackled by two stages: in the first stage, an off-the-shelf segmentation model, \eg, the Mask R-CNN~\cite{he2017mask} pretrained on COCO dataset, is used to extract the instance segmentation maps. Then, these object maps and visual features from the first stage are concatenated with audios, and fed into a PVT-v2 structure, to predict the final results. We denote this method as TwoSep, and the results are shown in Table~\ref{table:two_sep_baseline}.
It indicates that the audio-visual segmentation task is \textit{Not} bottlenecked by the first-stage segmentation quality, as the final performance is almost unchanged if using a much stronger Mask R-CNN (backbone from ResNet50 to powerful ResNeXt101), \eg, $\mathcal{M}_{\mathcal{J}}$ is 0.503 \textit{vs.} 0.502 in MS3 setting. Instead, without or with audios would largely affect the performance, \eg, $\mathcal{M}_{\mathcal{J}}$ is 0.473 \textit{vs.} 0.503. 
This again reflects the positive impact of audio signals, especially in the MS3 setting.
Our proposed framework consistently outperforms this baseline by a large margin.

\begin{table}[t]
\renewcommand\arraystretch{.6}
\footnotesize
\caption{\textbf{Comparison with a two-stage baseline method (\textbf{TwoSep})} which first generates the instance segmentation maps by the off-the-shelf Mask R-CNN and then combines audio signal for final sounding objects segmentation.
The performance is not bottlenecked by the segmentation quality (with different  Mask-RCNN backbones) but is largely influenced by the audio signal. Even using audio, our AVS method is much superior than this two-stage method.}
\begin{center}
 \begin{threeparttable}
  \begin{tabular}{   p{1.0cm}p{1.0cm}<{\centering}p{1.3cm}<{\centering}p{1.8cm}<{\centering}p{1.3cm}<{\centering}p{1.8cm}<{\centering}p{1.8cm}<{\centering}}
  \toprule[0.8pt]
      \multirow{3}{*}{Metric} & \multirow{3}{*}{Setting} & \multicolumn{2}{c}{TwoSep wo. audio}  & \multicolumn{2}{c}{TwoSep w. audio}    & AVS\\ 
    \cmidrule(r){3-4}\cmidrule(r){5-6}\cmidrule(r){7-7} 
                   & & {Res50} & {ResNeXt101} & {Res50} & {ResNeXt101} & {-} \\ \midrule
     \multirow{2}{*}{$\mathcal{M}_{\mathcal{J}}$} & S4  & .696 & .670 & .717 & .718 & \textbf{.787}  \\
         \cmidrule(r){3-7}
    & MS3     & .473  & .474 & .503 & .502 & \textbf{.540}  \\ 
    \bottomrule[0.8pt]
  \end{tabular}
\end{threeparttable}
\end{center}\label{table:two_sep_baseline}
\end{table}

\subsection{Analysis of the core components}\label{sec:exp_AVS_study}
\noindent\textbf{Impact of audio signal and TPAVI}.
As illustrated in Fig.~\ref{fig:TPAVI_module}, the TPAVI module is used to formulate the audio-visual interactions from a temporal and pixel-wise level, introducing the audio information to explore the visual segmentation.
We conduct an ablation study to explore its impact as shown in Table~\ref{table:w_wo_tapvi}.
Two rows show the proposed AVS method with or without the TPAVI module, while {``A$\oplus$V''} indicates directly adding the audio to visual features.
It will be noticed that adding the audio features to the visual ones does not result in a clear difference under the S4 setting, but lead to a distinct gain under the MS3 setting.
This is consistent with our hypothesis that audio is especially beneficial to samples with multiple sound sources,
because the audio signals can guide which object(s) to segment.
Furthermore, with the power of our TPAVI module, we can achieve a temporal and pixel-wise mapping.
With TPAVI, each visual pixel hears the current sound and the sounds at other times, while simultaneously interacting with other pixels.
The physical interpretation is that the pixels with high similarity to the same sound are more likely to belong to one object.
TPAVI helps further enhance the performance over various settings and backbones, \eg, $\mathcal{M}_{\mathcal{J}}$ is 0.728 \textit{vs.} 0.705 when using ResNet50 as the backbone under the S4 setting, and 0.531 \textit{vs.} 0.516 if using PVT-v2 under the MS3 setting.
Additionally, it is worth noting that the convolution blocks in the TPAVI module allow to project the input visual and audio features to the latent spaces that are suitable for attention computation.
For example, under the S4 setting and using ResNet50 as the backbone, if abandoning the four convolution blocks in
the TPAVI module, the $\mathcal{M}_{\mathcal{J}}$ will significantly drop from
0.728 to 0.592.

\begin{table}[t]
\caption{\textbf{Impact of audio signal and TPAVI.} Results  ($\mathcal{M}_{\mathcal{J}}$) of AVS both with and without the TPAVI module.  The middle row indicates directly adding the audio and visual features, which already improves performance under the MS3 setting.  The TPAVI module further enhances the results over all settings and backbones. }
\small
\begin{center}
\begin{threeparttable}
\setlength{\tabcolsep}{8pt}
\begin{tabular}{lcccc}
   \toprule[0.8pt]
       \multirow{2}{*}{AVS method}  & \multicolumn{2}{c}{S4}  & \multicolumn{2}{c}{MS3} \\ 
    \cmidrule(r){2-3}\cmidrule(r){4-5} 
                  & ResNet50 & PVT-v2   & ResNet50 & PVT-v2 \\ \midrule
    without TPAVI & .701    & .778  & .436    & .482 \\
    with A$\oplus$V &  .705 & .777 & .457 & .516 \\
    with TPAVI & \textbf{.728} & \textbf{.787} &  \textbf{.466} & \textbf{.531} \\
    \bottomrule[0.8pt]
   \end{tabular}
\end{threeparttable}
\end{center}\label{table:w_wo_tapvi}

\end{table}

\begin{figure}[t]
\centering
\includegraphics[width=\textwidth]{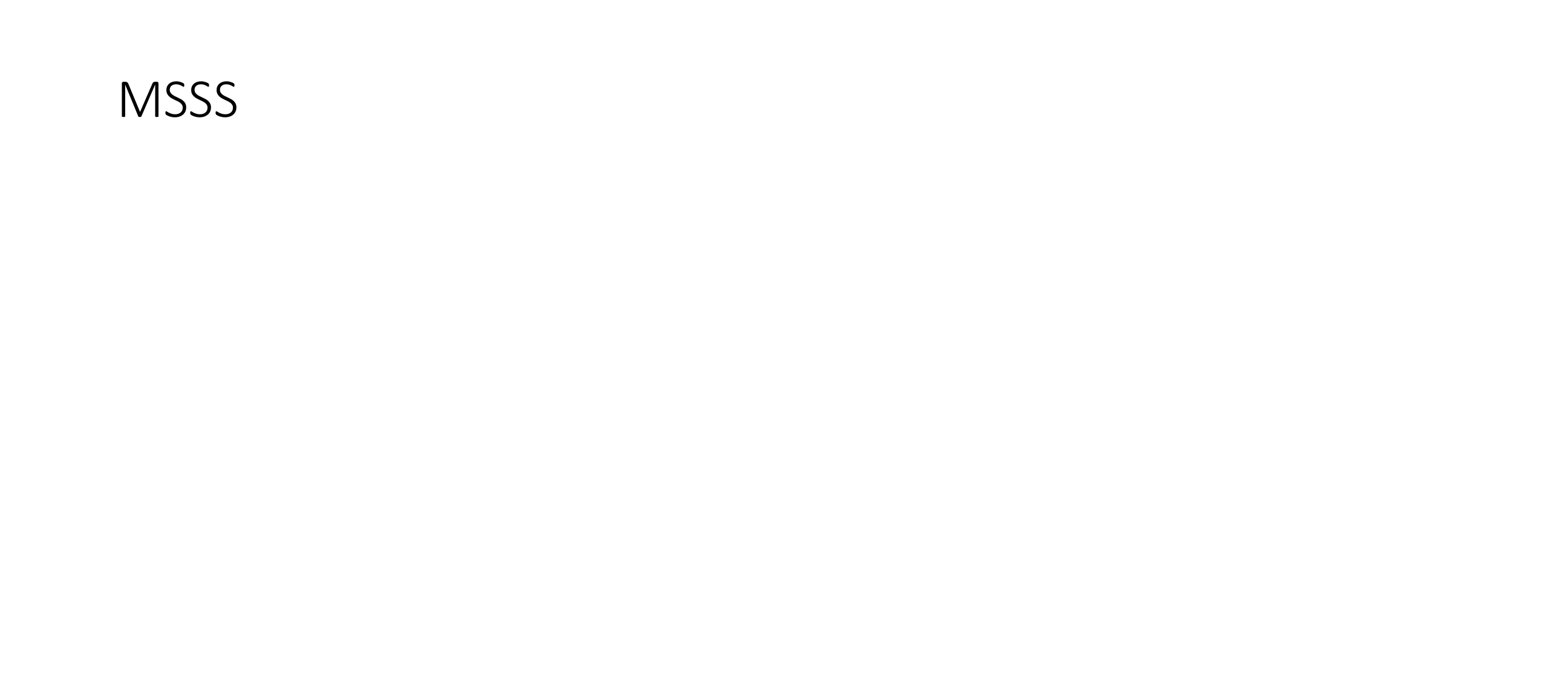}
\vspace{-6mm}
\caption{\textbf{Qualitative results under the semi-supervised S4 setting}. Predictions are generated by the ResNet50-based AVS model. Two benefits are noticed by introducing the audio signal (TPAVI): 1) learning the shape of the sounding object, \eg, \emph{guitar} in the video ({\sc Left}); 2) segmenting according to the correct sound source, \eg, the \emph{gun} rather than the \emph{man} ({\sc Right}).}
\label{fig:w_wo_tpavi_ssss}
\vspace{-4mm}
\end{figure}

\begin{figure}[t]
\centering
\includegraphics[width=\textwidth]{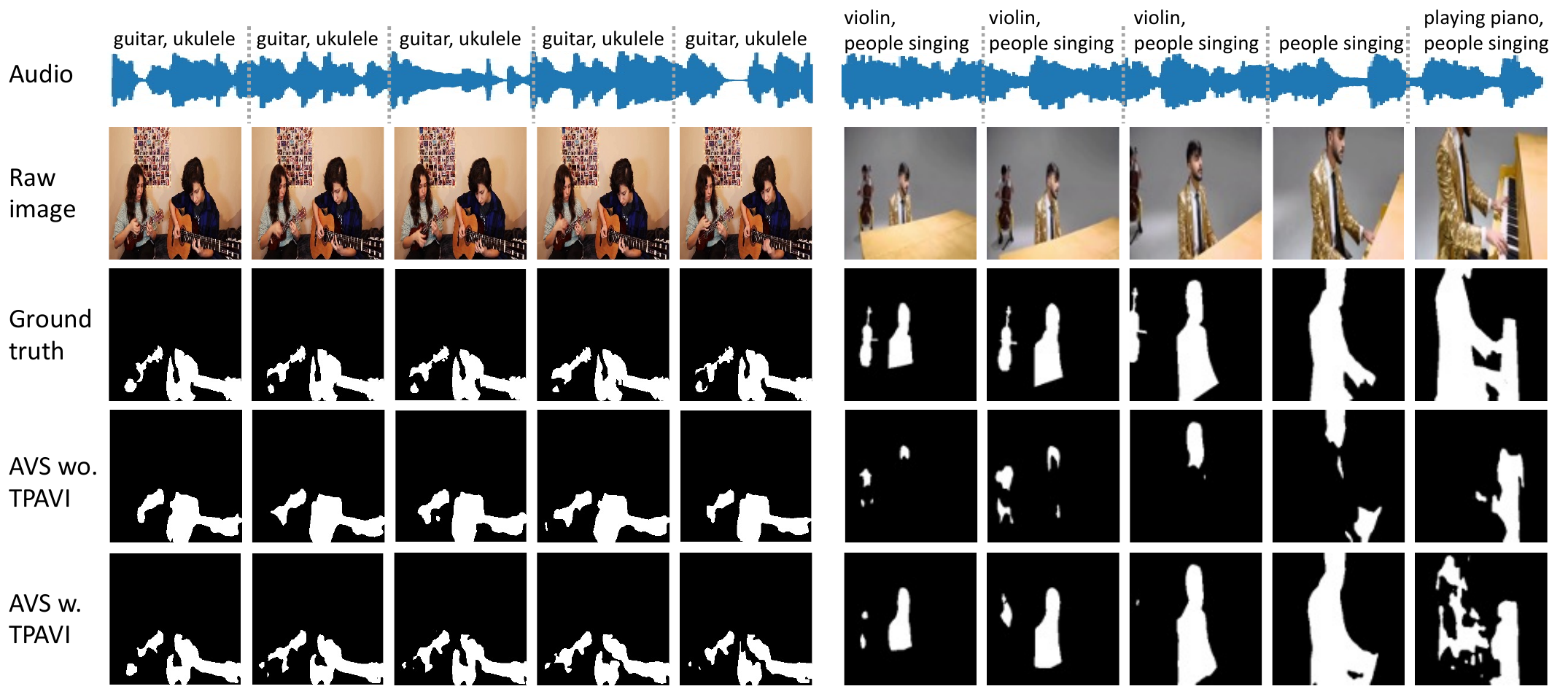}
\vspace{-6mm}
\caption{\textbf{Qualitative results under the fully-supervised MS3 setting}. The predictions are obtained by the PVT-v2 based AVS model. Note that AVS with TPAVI uses audio information to perform better in terms of 1) filtering out the distracting visual pixels that do not correspond to the audio, \ie, the \emph{human hands} ({\sc Left}); 2) segmenting the correct sound source in the visual frames that matches the audio more accurately, \ie, the \emph{singing person} ({\sc Right}).}
\label{fig:w_wo_tpavi_msss}
\end{figure}

We also visualize some qualitative examples to reflect the impact of  TPAVI. As shown in Fig.~\ref{fig:w_wo_tpavi_ssss}, the AVS method with TPAVI depicts the shape of sounding object better, \eg, the \emph{guitar} in the left video, while it can only segment several parts of the guitar without TPAVI. Such benefit can also be observed in MS3 setting, as shown in Fig.~\ref{fig:w_wo_tpavi_msss}, the model enables to ignore those pixels of \emph{human hands} with TPAVI.
More importantly, with TPAVI, the model is able to segment the correct sounding object and ignore the potential sound sources which actually do not make sounds, \eg, the \emph{man} on the right of Fig.~\ref{fig:w_wo_tpavi_ssss}.
Also, the ``AVS w. TPAVI'' has stronger ability to capture multiple sound sources. As shown on the right of Fig.~\ref{fig:w_wo_tpavi_msss}, the \emph{person} who is singing is mainly segmented with TPAVI but is almost lost without TPAVI. 
These results show the advantages of utilizing the audio signals, which helps to segment more accurate audio-visual semantic-corresponding pixels.

\begin{figure}[t]
\centering
\includegraphics[width=\textwidth]{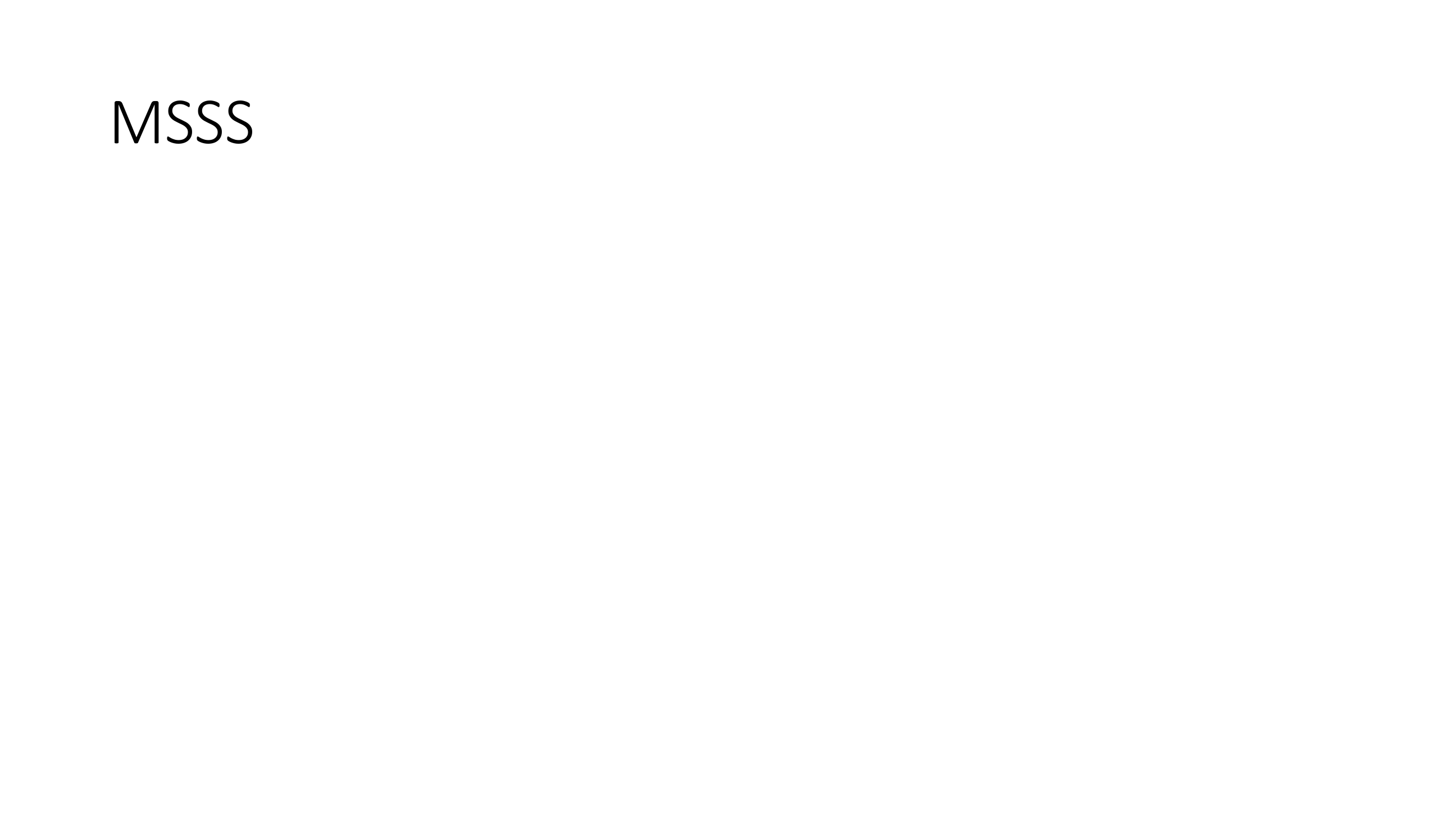}
\vspace{-4mm}
\caption{\textbf{Audio-Visual attention maps that come from the fourth stage TPAVI.} A brighter color indicates a higher response.
Such heatmaps are usually adopted as the final results for the SSL task, while they are just the intermediate output of the TPAVI module in our AVS framework. These results reveal that the TPAVI helps the model focus more on the visual regions that are semantic-corresponding to the audio.}
\label{fig:audio_visual_attention}
\vspace{3mm}
\end{figure}

Besides, we also visualize the audio-visual attention matrices to explore what happens in the cross-modal fusion process of TPAVI.
In detail, the attention matrix is obtained from $\alpha_i$ in Eq.~\eqref{eq:tpavi} of the fourth stage TPAVI.
We upsample it to have the same shape as the video frame. 
This is visually similar to the localization heatmap of these SSL methods, but only the intermediate result in our AVS method.
As shown in Fig.~\ref{fig:audio_visual_attention}, the high response area basically overlaps the region of sounding objects.
It suggests that TPAVI builds a mapping from the visual pixels to the audio signals, which is semantically consistent.

\noindent\textbf{Effectiveness of $\mathcal{L}_\text{AVM}$}. We expect that constructing the mapping between audio and visual features will enhance the network's ability to identify the correct objects. 
Therefore, we propose a $\mathcal{L}_\text{AVM}$ loss to introduce a soft constraint for training.
We only apply $\mathcal{L}_\text{AVM}$ in the fully-supervised MS3 setting because the change of sounding objects only happens there.

As shown in Table~\ref{table:loss_avm_study}, we explore two variants of the $\mathcal{L}_\text{AVM}$ loss.
$\mathcal{L}_{\text{AVM-AV}}$ is the one introduced in Eq.~\eqref{eq:avmloss}.
It encourages the visual features masked by the segmentation result to be consistent with the corresponding audio features in a statistical way, \ie, both depicting the sounding objects. 
%
Alternatively, $\mathcal{L}_{\text{AVM-VV}}$ first finds the closest audio partner for each candidate audio, and then computes the KL distance of the corresponding visual features (also masked by the segmentation results).
This is based on the idea that if two clips share similar audio signals, the visual features of their sounding objects should also be similar.
As shown in Table~\ref{table:loss_avm_study}, both variants achieve a clear performance gain.
%
For example, $\mathcal{L}_{\text{AVM-AV}}$ improves the $\mathcal{M}_{\mathcal{J}}$ by around 1\% and $\mathcal{M}_{\mathcal{F}}$ by about 2\%.
%
This demonstrates the benefits of introducing such an audio-visual constraint.
We use $\mathcal{L}_{\text{AVM-AV}}$,
since  $\mathcal{L}_{\text{AVM-VV}}$ inconveniently requires a ranking operation.

\begin{table}[t]
\vspace{-3mm}
\caption{\textbf{Effectiveness of $\mathcal{L}_{\text{AVM}}$.} The two variants of $\mathcal{L}_{\text{AVM}}$ both bring a clear performance gain compared with only using a standard BCE loss.}
\begin{center}
\vspace{-2mm}
\begin{threeparttable}
\setlength{\tabcolsep}{8pt}
\begin{tabular}{lcccc}
   \toprule[0.8pt]
       \multirow{2}{*}{Objective function}  & \multicolumn{2}{c}{MS3 ($\mathcal{M}_{\mathcal{J}}$)} & \multicolumn{2}{c}{MS3 ($\mathcal{M}_{\mathcal{F}}$)}\\ 
    \cmidrule(r){2-3}\cmidrule(r){4-5}
                  & ResNet50 & PVT-v2  & ResNet50 & PVT-v2 \\ \midrule
    $\mathcal{L}_{\text{BCE}}$  & .466    & .531 & .558 & .626 \\
    $\mathcal{L}_{\text{BCE}}$ + $\mathcal{L}_{\text{AVM-VV}}$ & .467 & .538 & .577 & .644\\
    $\mathcal{L}_{\text{BCE}}$ + $\mathcal{L}_{\text{AVM-AV}}$ & \textbf{.479} & \textbf{.540} & \textbf{.578} & \textbf{.645}\\
    \bottomrule[0.8pt]
   \end{tabular}
\end{threeparttable}
\end{center}\label{table:loss_avm_study}
\vspace{-8mm}
\end{table}

\begin{table}[t]
\caption{\textbf{Cross-modal fusion at various stages, measured by $\mathcal{M}_{\mathcal{J}}$.} In both the S4 and MS3 settings, the model achieves the best performance when the TPAVI module is used in all four stages.}
\begin{center}
\begin{threeparttable}
  \begin{tabular}{llp{1.cm}<{\centering}p{1.2cm}<{\centering}p{1.2cm}<{\centering}p{1.2cm}<{\centering}|p{1.2cm}<{\centering}p{1.2cm}<{\centering}p{1.2cm}<{\centering}}
  \toprule[0.8pt]
      \multirow{2}{*}{Setting}  & \multirow{2}{*}{Backbone}  & \multicolumn{7}{c}{$i$-th stage of Encoder, $i\in\{1,2,3,4\}$} \\ 
    \cmidrule(r){3-9} 
      & & 1 & 2 & 3 & 4 & 3,4 & 2,3,4 & 1,2,3,4 \\ \midrule
    \multirow{2}{*}{S4} & ResNet50 & .686 & .696 & \textbf{.713} & .670 & .713 & .720 & \textbf{.728} \\ 
    & PVT-v2 & .783 & \textbf{.786} & .780 & .777 & .782 & .785 & \textbf{.787} \\ \midrule
    \multirow{2}{*}{MS3} & ResNet50 & .416 & .424 & \textbf{.430} & .423 & .448 & .460 & \textbf{.479} \\
    & PVT-v2 & .462 & .488 & .474 & \textbf{.490} & .498 & .505 & \textbf{.540} \\
    \bottomrule[0.8pt]
  \end{tabular}
\end{threeparttable}
\end{center}\label{table:tpavi_at_different_stages}
\end{table}

\noindent\textbf{Cross-modal fusion at various stages}.
The TPAVI module is a plug-in architecture that can be applied in any stage for cross-modal fusion.
As shown in Table~\ref{table:tpavi_at_different_stages}, when the TPAVI module is used in different single stage, the segmentation performance fluctuates.
For the variant based on the ResNet50 backbone, the model achieves the best performance when employing the TPAVI module at the third stage under both S4 and MS3 settings. 
As for the PVT-v2 based model, it is better to use the TPAVI module at the second or fourth stages in the S4 and MS3 settings, respectively.
We attribute it to that the visual features at the first stage enjoy limited semantics.
Since our decoder architecture adopts a skip-connection, it would be beneficial to apply the TPAVI modules in multiple stages, as verified in the right part of Table~\ref{table:tpavi_at_different_stages}.
For example, under the MS3 setting, applying the TPAVI modules at all the four stages would increase the metric $\mathcal{M}_{\mathcal{J}}$ from  $0.490$ to $0.540$, with a gain of $5\%$.
It indicates the model has the ability to fuse and balance the features from multiple stages.

\begin{figure}[t]
\centering
\vspace{-2mm}
\includegraphics[width=\textwidth]{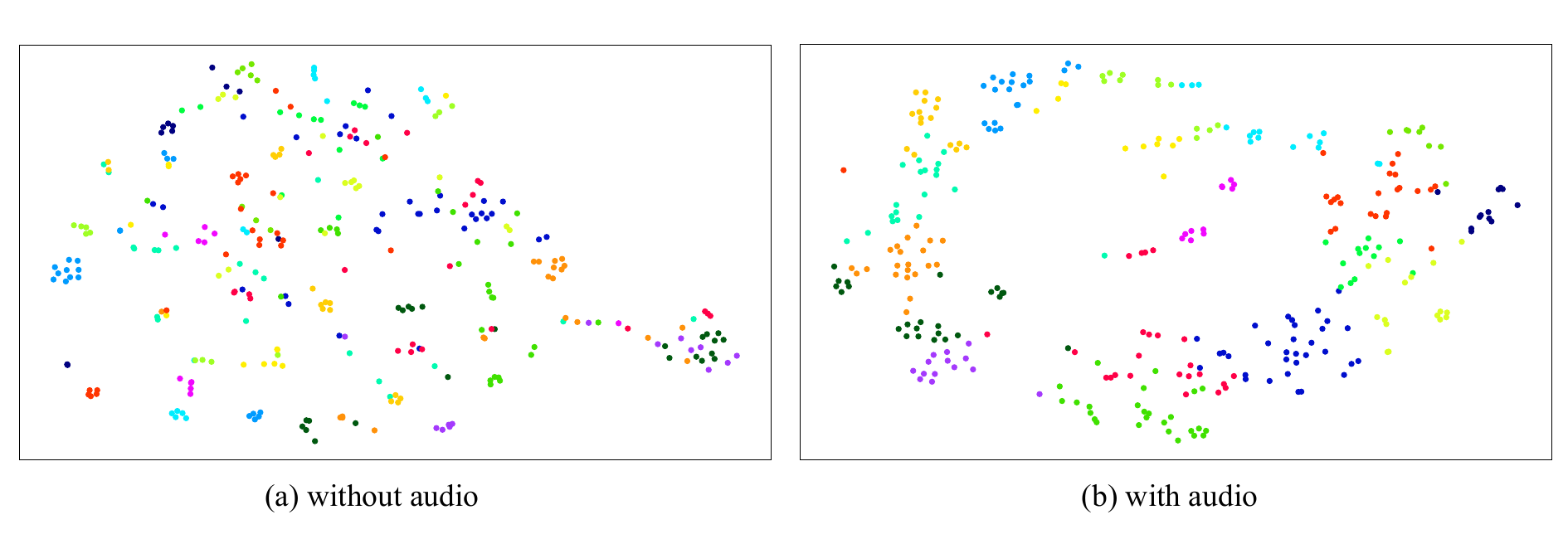}
\vspace{-6mm}
\caption{\textbf{T-SNE~\cite{tsne} visualization of the visual features, trained with or without audio.} These results are from the test split of the Multi-sources subset. We first use principal component analysis (PCA) to divide the audio features into $K=20$ clusters. Then we assign the audio cluster labels to the corresponding visual features and conduct t-SNE visualization. The points with the same color share the same audio cluster labels. It can be seen that when training is accompanied by audio signals (right), the visual features illustrate a closer trend with the audio feature distribution, \ie, points with the same colors gather together, which indicates an audio-visual correlation has been learned.  (Best viewed in color.) }
\label{fig:tsne}
\vspace{-4mm}
\end{figure}

\noindent\textbf{T-SNE visualization analysis.}
We also visualize the visual features with or without TPAVI module to analyze whether the network has built a connection between the audio and the visual features. 
Specifically, on the test split of the Multi-sources set, we use the PVT-v2 based AVS model to obtain the visual features.
Since the Multi-source set do not have category labels (its videos may contain several categories), 
we use the principal component analysis (PCA) to divide the audio features into $K=20$ clusters. Then we assign the audio cluster labels to the corresponding visual features. In this case, if the audio and the visual features are correlated, the visual features should be clustered as well. We use the t-SNE visualization to verify this assumption.  
As shown in Fig.~\ref{fig:tsne}a, without audio signals, the learned visual features distribute chaotically;
whereas in Fig.~\ref{fig:tsne}b, the visual features sharing the same audio labels tend to gather together.
This indicates that the distribution of the visual features and  audio features are highly correlated.

\begin{table}[t]
\caption{\textbf{Performance with different initialization strategies under the MS3 setting.}
Compared to training from scratch under the MS3 setting, we observe a significant performance improvement if pre-training the model on the Single-source subset. Note the proposed $\mathcal{L}_{\text{AVM}}$ loss is used in all the experiments of the Table. The metric is $\mathcal{M}_{\mathcal{J}}$.}
\begin{center}
\begin{threeparttable}
  \begin{tabular}{lp{2cm}<{\centering}p{2cm}<{\centering}p{2cm}<{\centering}p{2cm}<{\centering}}
  \toprule[0.8pt]
      \multirow{2}{*}{AVS method}  & \multicolumn{2}{c}{ From scratch}  & \multicolumn{2}{c}{ Pretrained on Single-source} \\ 
    \cmidrule(r){2-3}\cmidrule(r){4-5} 
                  & ResNet50 & PVT-v2   & ResNet50 & PVT-v2 \\ \midrule
    wo. TPAVI & .436    & .482 & \textbf{.455}    & \textbf{.506} \\
    w. TPAVI  & .479    & .540 & \textbf{.543}    & \textbf{.573} \\
    \bottomrule[0.8pt]
  \end{tabular}
\end{threeparttable}
\end{center}\label{table:load_ssss_for_msss}
\end{table}

\subsection{Discussion of the model training and inference}

\noindent\textbf{Without backbone pre-training.} We try to train the AVS framework without the pretrained backbones. As expected, we observe an obvious performance drop, \eg, the $\mathcal{M}_{\mathcal{J}}$ decreases from 0.728 to 0.441 with ResNet50 backbone under S4 setting. We speculate that it is difficult for the model to learn the audio and visual representation totally from scratch, especially for this challenging pixel-wise segmentation task. For a fair comparison, all the other experiments in the paper use the pretrained backbones. 

\noindent\textbf{Pre-training on the Single-source subset.}

As introduced in Sec.~\ref{sec:dataset} of the paper, the videos in the Multi-sources subset share similar categories to those in the Single-source subset.
A natural idea is whether we can pre-train the model on the Single-source subset to help deal with the MS3 problem.
As shown in Table~\ref{table:load_ssss_for_msss}, we test two initialization strategies, \ie, from scratch or pretrained on the Single-source subset.
It is verified that the pre-training strategy is beneficial in all the settings, whether we use the audio information (``AVS w. TPAVI'') or not (``AVS wo. TPAVI'').
Taking the PVT-v2 based AVS model for example, the $\mathcal{M}_{\mathcal{J}}$ is improved from $0.482$ to $0.506$ (by $2.4\%$) and from $0.540$ to $0.573$ (by $3.3\%$), respectively without or with TPAVI.
The phenomenon is more obvious if using ResNet50 as the backbone and adopting the TPAVI module, where the $\mathcal{M}_{\mathcal{J}}$ increases from $0.479$ to $0.543$ (by $6.4\%$).
With pre-training on the Single-source subset, the model can learn prior knowledge about the audio-visual correspondence, \ie, the matching relationship between the visual objects and sounds.
This kind of knowledge is naturally beneficial.

\begin{figure}[t]
\centering
\includegraphics[width=\textwidth]{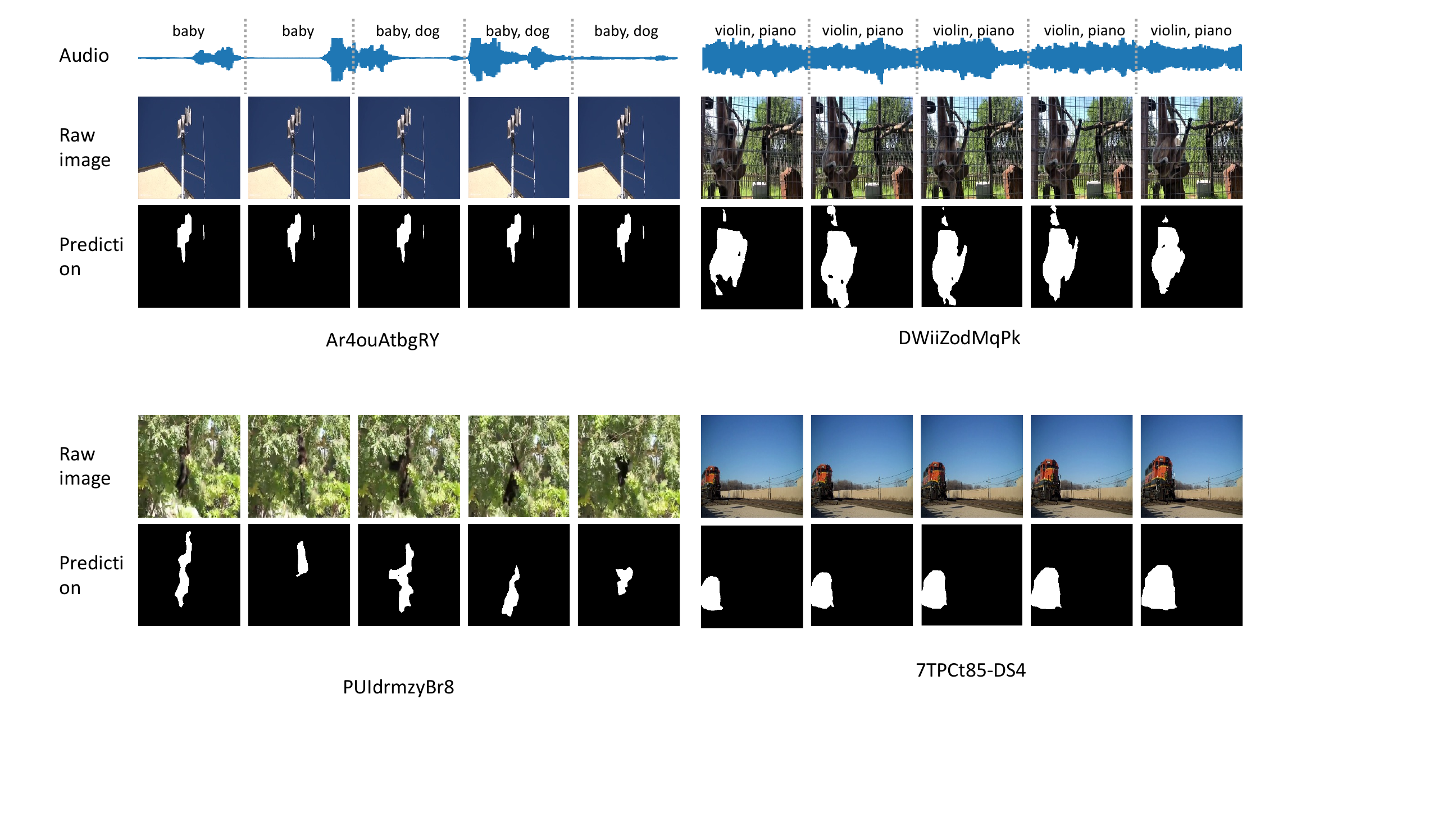}
\caption{\textbf{Qualitative examples of applying the pretrained AVS model to unseen videos.} The caption in each sub-figure indicates the sounding object(s) accordingly. There are almost no videos having the same category as these sounding objects during AVS model training. The pretrained AVS model gains the ability to segment the correct sounding object(s) in both single and multi sources.}
\label{fig:unseen_objects_inference}
\end{figure}

\noindent\textbf{Segmenting unseen objects.} The proposed AVS framework is trained without accessing the sounding object category labels, and hence it can be used to predict the videos which do not strictly lie in the category vocabulary of AVSBench dataset, though may have a performance drop for unseen objects. We display some qualitative visualisations on real-world videos whereas the category of sounding objects are barely not appeared in the training set of AVS model.
As shown in Fig.~\ref{fig:unseen_objects_inference}, the pretrained AVS model has a certain ability to segment the correct sounding objects in the case of single sound source (a), multiple visible objects (b, c), and multiple sound sources (d).
We speculate that the pretrained AVS model learned some prior knowledge about audio-visual correspondence from AVSBench dataset that helps it to generalize to even unseen videos and give possibly accurate pixel-level segmentation.
Although there may be some failures, the AVS model can be further strengthened if trained with videos of more sound source categories.

\section{Conclusion}
We have proposed a new task called AVS, which aims to generate pixel-level binary segmentation masks for sounding objects in audible videos. To facilitate research in this area, we collected the first audio-visual segmentation benchmark (called AVSBench). Depending on the number of sounding objects, two settings of AVS were explored: semi-supervised single-source (S4) and fully-supervised multi-source (MS3). We presented a new pixel-level AVS method to serve as a strong baseline, which includes a TPAVI module to encode the pixel-wise audio-visual interactions within temporal video sequences and a regularization loss to help the model learn audio-visual correlations. We compared our method with several existing state-of-the-art methods of the related tasks on AVSBench dataset, and further demonstrated that our method can build a connection between the sound and the appearance of an object. For future work, we believe this research will facilitate multimodal semantic segmentation, \ie, assigning semantic labels to both audio and visual segments in audible videos.

\vspace{5mm}
\noindent\textbf{Acknowledgement}
The research of Jinxing Zhou, Dan Guo, and Meng Wang was supported by
the National Key Research and Development Program of China (2018YFB0804205),
and the National Natural Science Foundation of China (72188101, 61725203).
Thanks to the SenseTime Research
for providing access to the GPUs used for conducting experiments.
The article solely reflects the opinions and conclusions of its authors but not the funding agents.

%
%
\bibliographystyle{splncs04}
\bibliography{egbib}

\end{sloppypar}
\end{document}